\def\year{2022}\relax
\documentclass[letterpaper]{article} 
\usepackage{aaai22}  
\usepackage{times}  
\usepackage{helvet}  
\usepackage{courier}  
\usepackage[hyphens]{url}  
\usepackage{graphicx} 
\usepackage{natbib}  
\usepackage{caption} 
\DeclareCaptionStyle{ruled}{labelfont=normalfont,labelsep=colon,strut=off} 
\usepackage{algorithm}
\usepackage{algorithmic}

\usepackage{slashbox}
\usepackage{multirow}

\usepackage{amsthm,amsmath}
\usepackage{amssymb}
\usepackage{mathrsfs}
\usepackage{bm}
\usepackage{bbm}

\usepackage{booktabs}

\usepackage{array}

\usepackage{listings}
\floatstyle{ruled}
\newfloat{listing}{tb}{lst}{}
\floatname{listing}{Listing}
\title{Generalizable Person Re-Identification via \\Self-Supervised Batch Norm Test-Time Adaption}
\author {
    Ke Han\textsuperscript{\rm 1,2},
    Chenyang Si\textsuperscript{\rm 1},
    Yan Huang\textsuperscript{\rm 1,3}\footnote{Corresponding author.},
    Liang Wang\textsuperscript{\rm 1,3,4,5},
    Tieniu Tan\textsuperscript{\rm 1,3,4}
}
\affiliations {
    \textsuperscript{\rm 1} Center for Research on Intelligent Perception and Computing,
Institute of Automation, Chinese Academy of Sciences \\
    \textsuperscript{\rm 2} School of Future Technology,
University of Chinese Academy of Sciences (UCAS)\\
    \textsuperscript{\rm 3} School of Artificial Intelligence,
University of Chinese Academy of Sciences (UCAS)\\
\textsuperscript{\rm 4} Center for Excellence in Brain Science and Intelligence Technology (CEBSIT) \\
\textsuperscript{\rm 5}
Chinese Academy of Sciences, Artificial Intelligence Research (CAS-AIR) \\
    ke.han@cripac.ia.ac.cn, chenyang.si.mail@gmail.com, \{yhuang, wangliang, tnt\}@nlpr.ia.ac.cn
}

\begin{document}

\maketitle

\begin{abstract}
In this paper, we investigate the generalization problem of person re-identification (re-id), whose major challenge is the distribution shift on an unseen domain.
As an important tool of regularizing the distribution, batch normalization (BN) has been widely used in existing methods.
However, they neglect that BN is severely biased to the training domain and inevitably suffers the performance drop if directly generalized without being updated.
To tackle this issue, we propose Batch Norm Test-time Adaption (BNTA), a novel re-id framework that applies the self-supervised strategy to update BN parameters adaptively.
Specifically, BNTA quickly explores the domain-aware information within unlabeled target data before inference, and accordingly modulates the feature distribution normalized by BN to adapt to the target domain.
This is accomplished by two designed self-supervised auxiliary tasks, namely part positioning and part nearest neighbor matching, which help the model mine the domain-aware information with respect to the structure and identity of body parts, respectively.
To demonstrate the effectiveness of our method, we conduct extensive experiments on three re-id datasets and confirm the superior performance to the state-of-the-art methods.

\end{abstract}

\begin{figure}[!h]
\centering
\includegraphics[width=\columnwidth]{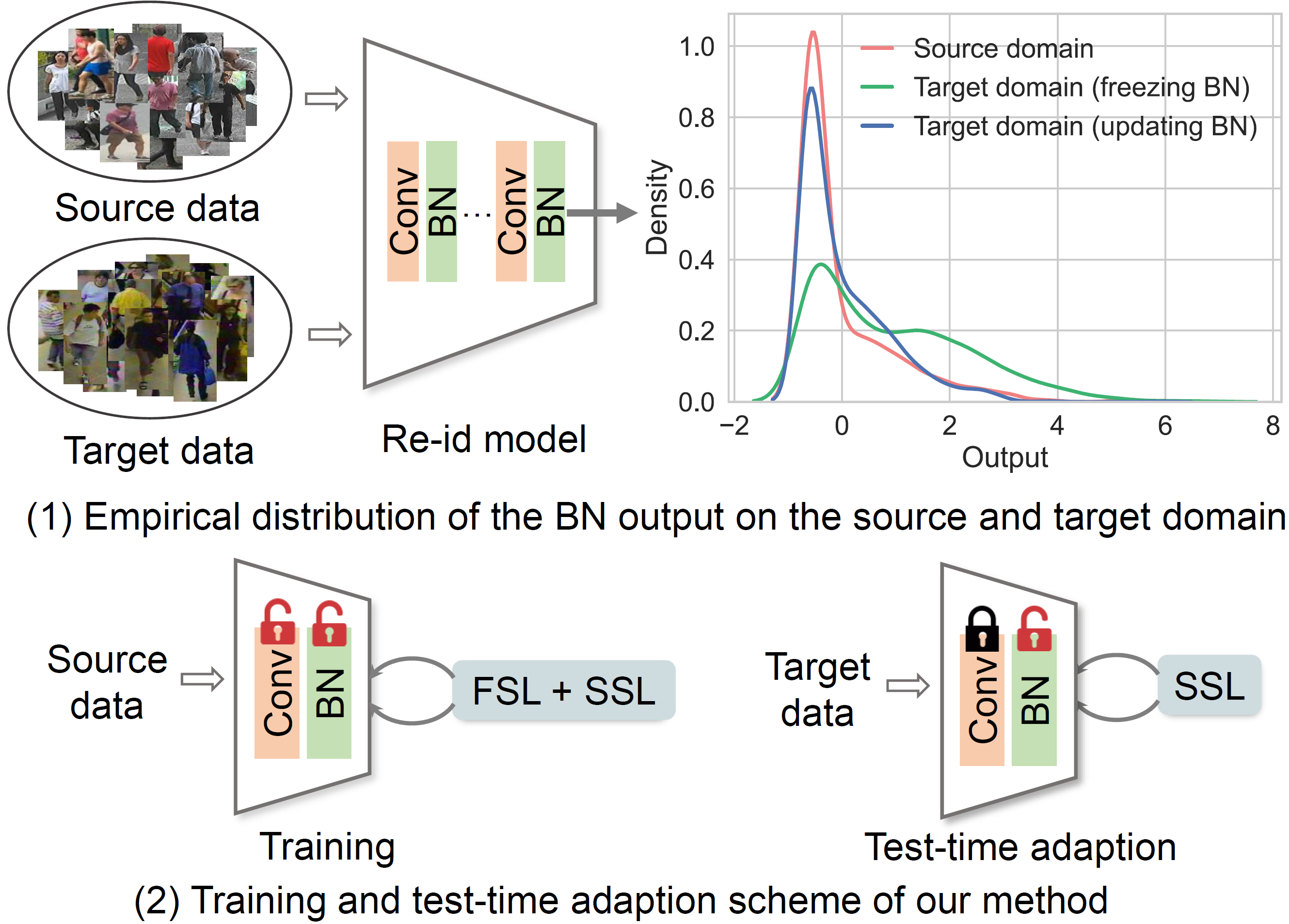} 
\caption{
Illustration of our main idea. (1) The empirical distribution of the BN outputs shifts largely from the source to the target domain when BN layers are frozen, but the shift is mitigated when BN layers are updated.
(2) Our model is jointly optimized via FSL and SSL heads during training. BN layers are updated via SSL, while other layers remain frozen during test-time adaption.}
\label{intro}
\vspace{-5mm}
\end{figure}

\vspace{-2mm}
\section{Introduction}
Person re-identification (re-id) aims to match individuals with the same identity across cameras at different locations over a large disjoint space.
A number of efforts have been made by the re-id community over the past years, making remarkable progress in scenarios where training and test data come from the same domain.
In reality, however, if tested directly on a previously unseen domain, most existing methods suffer significant performance degradation due to the distribution shift in background \cite{mask-reid}, resolution \cite{han_2}, clothing styles \cite{hy_1}, etc.
Improving the generalization ability is therefore greatly important for promoting the study of re-id. 


Recently, generalizable re-id methods have drawn increasing attention,
which can be roughly divided into three categories.
The methods of the first category, based on meta-learning, mimic the train-test splits to improve the ability of dealing with the stimulated generalization situations \cite{dmg-net,metabin}.
The second methods aim to learn domain-invariant re-id features by the memory mechanism \cite{dimn}, hard example mining \cite{augming} or adversarial learning \cite{ddan}.
The third methods 
advance the common usage of batch normalization (BN), \emph{e.g.}, combining it with instance normalization (IN) to offset domain-related information captured by BN \cite{dualnorm,SNR}.

The third methods reveal that BN discourages the generalization ability of the model, due to learning the knowledge biased to the source domain.
We further explicitly investigate the correlation between distributions of BN outputs and domains in Fig. \ref{intro}. (1). The red and green lines indicate the empirical Gaussian-like distributions output by the same (freezing) BN layer on the source and target domain, respectively.
They exhibit a great distribution shift, \emph{e.g.}, the variance increases obviously from the red line to the green one, suggesting the BN outputs on the target domain are dispersed more widely.
The reason lies in that BN parameters are severely biased to the training data when regularizing the distribution, but the target data comes from a quite different distribution.
This leads to two consequences.
1) The input distribution to the following layers (\emph{e.g.}, convolutional layers) is accordingly deviated from that on the source domain,
which affects adversely their accuracy of handling information.
2) The shift has even been accumulated for the top layers, thereby weakening the discriminativeness of the output features.
However, this issue is not tackled well by existing methods, subject to their applying the trained BN layers to an unseen distribution directly without any updating.


Target samples, despite without identity labels, carry underlying prior information about the target distribution. 
It can be exploited to correct the training bias stored in BN layers for adapting to the target domain before inference, thus mitigating the distribution shift and enhancing the generalization performance. 
However, how to quickly explore the domain-aware information from unlabeled target data is highly challenging and rarely investigated in re-id.
Self-supervised learning (SSL) has recently been proven very effective in unsupervised learning in the classification task, such as MoCo \cite{moco} and BYOL \cite{byol}.
Experimentally, they do not suit to be directly applied to re-id because of the task gap, making it urgent to design re-id oriented SSL tasks for unsupervised learning.

In this paper, we propose a novel Batch Norm Test-time Adaption (BNTA) re-id framework to update BN layers via self-supervision.
Specifically, BNTA fast explores the domain-aware information within unlabeled target samples, and accordingly updates BN parameters (including statistics and affine parameters) to modulate the normalized feature distribution on the target domain.
Inspired by the previous works verifying both body structure and identity information are important to re-id \cite{auto_reid}, we design two SSL auxiliary tasks for re-id named part positioning and part nearest neighbor matching.
They help the model mine the target distribution involved with the structure and identity cues of body parts, by predicting the position and exploiting the similarity between the nearest neighbors for body parts, respectively.
Furthermore, based on the two SSL tasks, we present a training and test-time adaption scheme.
As illustrated in Fig \ref{intro}. (2), our model is trained jointly via FSL (fully-supervised learning) and SSL on the labeled source data, while SSL further enables updating BN layers during test-time adaption to absorb the target distribution.

As a result, as shown in Fig \ref{intro}. (1), 
the outputs of the updated BN layer on the target domain (blue line) have a similar distribution to that on the source domain (red line), which ensures that the following layers receive a stable input distribution and effectively enhances the generalization performance.
Extensive experiments on VIPeR, GRID and iLIDS datasets demonstrate the effectiveness of our method, and confirm the advantage over the state-of-the-art methods.
Besides, the proposed test-time adaption is fast and easy to implement, which only takes a few seconds with hundreds of gallery samples without additional target data collection.
The generalization performance is even further boosted as the number of target samples increases, showing the potential of our model in the real-world scenarios where plenty of unlabeled target images are generally available.

We summarize the contributions of this paper as follows.
\begin{itemize}
\item To alleviate the distribution shift when transferring BN layers to an unseen domain, we propose a BNTA re-id framework for fast updating test-time BN parameters. 
\item Two simple yet effective SSL auxiliary tasks are designed to explore the structure and identity information of body parts from the unlabeled target data for BNTA.
\item Extensive experiments demonstrate the state-of-the-art performance of our model on three re-id datasets, and also promote the understanding about how and why updating BN parameters improves the re-id generalization.
\end{itemize}

\vspace{-2mm}

\section{Related Work}
\textbf{Cross-Domain Person Re-Identification.}
Cross-domain re-id addresses the re-id performance drop across domains, assuming that data from the labeled source domain and unlabeled target domain are both utilizable during training.
Mainstream methods 
include transferring the image style from the source to target domain \cite{uda_reid_1,uda_reid_2}, clustering and generating pseudo labels \cite{hy_4,uda_reid_cluster1}, and learning domain-aligned features \cite{uda_reid_align1,hy_3,hy_2,nk_3}.
However, they require large-scale target data collection (generally at least thousands of target images) 
and plenty of training iterations, which is highly inflexible and time-consuming when deploying a re-id system to a new domain.
In contrast, our method does not need target data during training, and only takes a few seconds with hundreds of gallery images for the test-time adaption.

\begin{figure*}[t]
\centering
\includegraphics[width=0.96\textwidth]{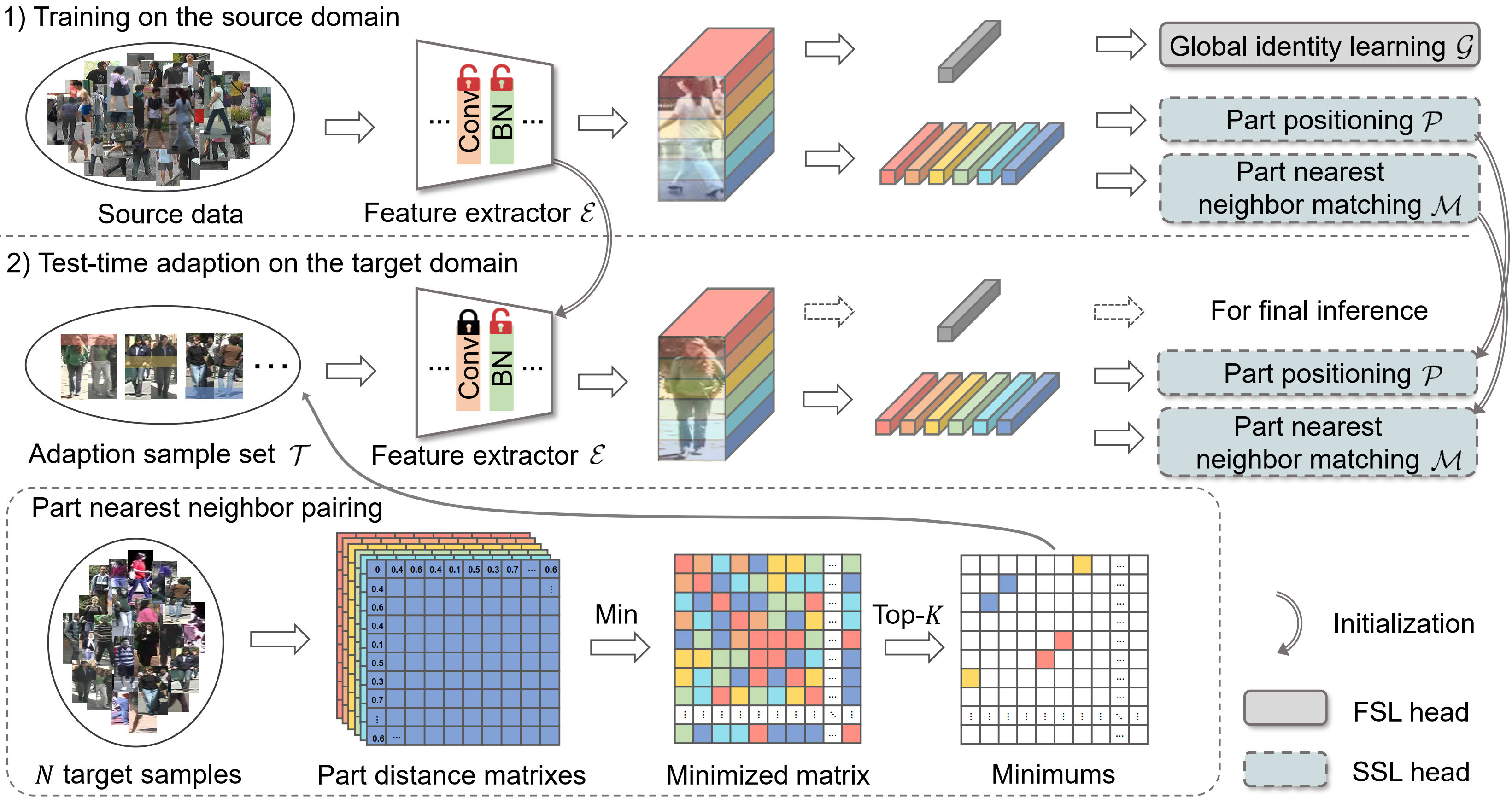} 
\caption{Illustration of the BNTA re-id framework. A FSL head $\mathcal{G}$ and two auxiliary SSL heads $\mathcal{P},\mathcal{M}$ enable the training of the whole model on the labeled source domain based on global and local features, respectively. During test-time adaption, BN layers are updated via SSL using the adaption sample set, which includes $K$ image pairs with part nearest neighbors selected from $N$ target samples.
The updated model extracts global features for the final inference.}
\label{method}
\vspace{-2mm}
\end{figure*}

\noindent
\textbf{Generalizable Person Re-Identification.}
Unlike cross-domain re-id, generalizable re-id aims to improve the generalization performance on unseen domains, which supposes the target data is invisible during training.
The main literature can roughly fall into three categories, \emph{i.e.}, meta-learning, domain-invariant learning and BN based methods. The representative works are introduced as follows, respectively.
\textbf{1)} Inspired by meta-learning, MetaBIN \cite{metabin} stimulates and learns to handle the unsuccessful generalization situations, whereas DMG-Net \cite{dmg-net} is a dual-meta network to exploit the meta-learning more fully in both the training procedure and metric space learning.
\textbf{2)} To learn domain-invariant representations, DIMN \cite{dimn} maps an image to its identity classifier with a novel memory bank module; AugMing \cite{augming} advances the strategy of data augmentation and selects the hard examples for increasing the utilization of data; DDAN \cite{ddan} selectively aligns the distribution of multiple source domains via domain-wise adversarial learning and identity-wise similarity enhancement.
\textbf{3)} Some methods, realizing the limitation of the common BN usage on the generalization ability, additionally combine IN to reduce the domain bias captured by BN with instance-related information \cite{dualnorm}, while restituting the identity-relevant information \cite{SNR}.

All above methods can only be trained on the training data, and applied to a new domain without being updated.
Different from them, our model has a domain-adaptive learnability, which can perceive the target distribution to adaptively update the model for re-regularizing the feature distribution and enhancing the generalization performance.

\noindent
\textbf{Batch Normalization.}
Since originally designed to stabilize the neural responses and training procedure, BN \cite{BN} has been used widely in deep neural networks.
However, BN layers are inclined to normalize the training data to some specific Gaussian-like distributions, and if the target data deviates from that dramatically, the model will generalize poorly.
To this end, CBN \cite{cbn} optimizes BN layers with independent statistics on each training domain, and then recomputes the statistics on the target domain like AdaBN \cite{adaBN}.
Also, some works \cite{BN_1,BN_2} make a step forward and learn both domain-specific statistics and affine parameters, considering the regulatory effect of affine parameters is also closely tied to the domain.
However, their learnable affine parameters are not updatable on the unlabeled target domain. 
Unlike them, our proposed SSL strategy can explore the domain-aware information for orienting BN parameters (both statistics and affine parameters) to the target distribution adaptively.


\section{Method}

\subsection{Overview}
In this paper, we propose a Batch Norm Test-time Adaption (BNTA) framework to improve the generalization ability for re-id, through updating test-time BN parameters via self-supervision.
The overview is illustrated in Fig. \ref{method}.


We begin the detailed methodology with defining some notations.
Following the generalizable re-id setting \cite{dualnorm}, the training set $\mathcal{D}$ is a mixture of $S$ source domains 
$\mathcal{D} =\left\{ \mathcal{D}_1, \mathcal{D}_2, \cdots, \mathcal{D}_s \right\}$.
Each domain is composed of labeled image pairs,
and $M_i$ is the number of training identities in $\mathcal{D}_i$.
Since $S$ source label sets are non-overlapping, there are totally $M=\sum_{i}M_i$ training identities in $\mathcal{D}$.
The test set is collected from a target domain that is different from all the source domains.

\subsection{FSL: Global Identity Learning}

The FSL head $\mathcal{G}$ aims to learn discriminative identity features from the source labeled data for re-id.
Many re-id models containing BN layers \cite{han_1,nk_1,nk_2} can be adopted as the backbone of our feature extractor $\mathcal{E}$. Here we choose DualNorm \cite{dualnorm} for its competitive generalization ability and relatively concise structure.
For a given input image $\bm{x}_i \in \mathcal{D}$, we denote the extracted feature maps as $\mathcal{E}(\bm{x}_i) \in \mathbb{R}^{C\times H\times W}$, where $C$ is the number of channels, $H$ and $W$ are the height and width, respectively.
The following operation for identity learning is expressed as
\begin{flalign}
    && &\bm{f}_i = {\rm Avgpool2d}(\mathcal{E}(\bm{x}_i)), &  \\
    && &\mathcal{L}_{id} = - \bm{y}_i \cdot {\rm log}({\rm FC_\mathcal{G}}({\bm{f}_i})), &
\label{id_loss}
\end{flalign}
where $\bm{f}_i$ is the identity feature vector after global average pooling with the dimension $C$, ${\rm FC_\mathcal{G}}(\cdot)$ is a fully-connected layer followed by a softmax function, and $\bm{y}_i \in \mathbbm{1}^{1 \times M}$ is the binary identity label.
The identity loss $\mathcal{L}_{id}$ is a cross-entropy loss for making identity features discriminative by identity classification.
\begin{equation}
\label{A_tri}
{\mathcal{L}^{tri} = \sum_{p=1}^P \max(0, \phi_{tri} + d(f_{v_i}^*,f_{v_i^{p+}}^*) - d(f_{v_i}^*,f_{v_i^{p-}}^*)),}
\end{equation}
where $d(\cdot,\cdot)$ is the Euclidean distance function. $\{f_{v_i}^*,f_{v_i^{p+}}^*,f_{v_i^{p-}}^*\}$ is a set of triplet training samples where $f_{v_i}^*$ is an anchor, $f_{v_i^{p+}}^*$ and $f_{v_i^{p-}}^*$ have the same and different identity label to $f_{v_i}^*$, respectively. $\phi_{tri}$ is the margin parameter, and $P$ is the number of the triplet sets for $v_i$ in a training batch.
This loss could pull intra-class feature distances closer and push inter-class feature distances away.


\subsection{SSL: Part Positioning}
The body structure information plays an important role in re-id but is easily neglected \cite{auto_reid}.
In addition, person images always have a clear structural prior, \emph{i.e.}, on a person image from top to bottom is always head to feet, even captured on different domains.
Inspired by the two findings, we present a SSL auxiliary head $\mathcal{P}$ named part positioning to explore the body structure within the images by predicting the positions of body parts.
Specifically,
\begin{flalign}
\label{eq_5}
    && \bm{f}_i^h & = {\rm Avgpool2d}_h(\mathcal{E}(\bm{x}_i)), &  \\
    && \mathcal{L}_{pos} & = - \sum_{h=1}^H \bm{y}_i^h \cdot {\rm log}({\rm FC}_\mathcal{P}(\bm{f}_i^h)). &
\label{pred_loss}
\end{flalign}
where $\left\{\bm{f}_i^h\right\}_{h=1}^H$ are local feature vectors obtained by dividing the feature map evenly and average pooling, corresponding to $H$ vertical body parts from top to bottom on $\bm{x}_i$.
They are sent into a fully-connected layer $\mathcal{P}$ all together to predict the vertical position indexes (1, 2, ..., or $H$), along with the given binary labels $\bm{y}_i^h \in \mathbbm{1}^{1 \times H}$,
which is supervised by the positioning loss $\mathcal{L}_{pos}$.
When applying our model to a new domain, $\mathcal{L}_{pos}$ can promote the model to reduce the feature distribution shift from the source domain by perceiving and aligning the structure information within images.

\subsection{SSL: Part Nearest Neighbor Matching}
Since re-id relies on identity characteristics for image retrieval, the inter-image identity similarity also has a significant impact on the feature distribution, apart from the body structure.  
We thereby design another SSL head $\mathcal{M}$, namely part nearest neighbor matching, to mine the identity distribution on the target domain based on the local similarity. 
The motivation of exploiting local instead of global similarity is that the local one has more reliability and potential for unlabeled target images.
For example, when two images contain the seemingly same black shirts, even with different identities, we can still exploit them to simulate local positive pairs to explore the underlying inter-image identity similarity.

\subsubsection{Training Version.}
During training, local features are required to be initialized to be discriminative, so that they can be used to modulate the identity distribution in place of global features during test-time adaption.
The process for the local identity learning is formulated as
\begin{flalign}
\label{eq_7}
    && & \bm{f}_i^h \leftarrow  {\rm Conv}_\mathcal{M}^h(\bm{f}_i^h), &  \\
    &&  \mathcal{L}_{mat}^{t} = & - \sum_{h=1}^H \bm{y}_i \cdot {\rm log}( {\rm FC}_\mathcal{M}^h(\bm{f}_i^h) ), &
\end{flalign}
where ${\rm Conv}_\mathcal{M}^h(\cdot)$ is the convolutional layer that has the kernel size of $1 \times 1$ and transforms the dimension of $\bm{f}_i^h$ from $C$ to $C_l$.
Similar to Eq. (\ref{id_loss}) formally, $\mathcal{L}_{mat}^{t}$ is the training version of the part nearest neighbor matching loss, 
and capacitates our model to extract discriminative local features. 

\begin{algorithm}[tb]
\caption{Part nearest neighbor pairing}
\label{alg1}
\textbf{Input}: The trained feature extractor $\mathcal{E}$, the trained part nearest matching head $\mathcal{M}$, $N$ gallery images, the hyper-parameter $K$.  \\
\textbf{Output}: The adaption sample set $\mathcal{T}$.

\begin{algorithmic}[1] 
\FOR {$i=1$ \textbf{to} $N$}
    \STATE Extract features of body parts $\left\{\bm{f}_i^h\right\}_{h=1}^H$ via Eq. (\ref{eq_5}) and Eq. (\ref{eq_7}). 
       \quad // $\bm{f}_i^h \in \mathbb{R}^{C_l}$
\ENDFOR
\STATE Construct feature matrixes $\left\{ \textbf{\emph{M}}_h\right\}_{h=1}^H$. \hspace{-2mm} \quad // $ \textbf{\emph{M}}_h \in \mathbb{R}^{N \times C_l}$
\STATE Calculate part distance matrixes
$\left\{\textbf{\emph{D}}_h\right\}_{h=1}^H$ via Euclidean distance, and concatenate them as $\textbf{\emph{D}}$. \hspace{-2mm} \quad // $\textbf{\emph{D}} \in \mathbb{R}^{H \times N \times N}$
\STATE Take the minimum among $H$ part distances:  $\textbf{\emph{D}} \leftarrow \min(\textbf{\emph{D}}, \dim=0)$. \hspace{+28mm} \quad // $\textbf{\emph{D}} \in \mathbb{R}^{N \times N}$
\STATE Construct $\mathcal{T}$, composed of $K$ non-overlapping image pairs with the minimum part distances in $\textbf{\emph{D}}$ and the corresponding position labels.
\end{algorithmic}
\end{algorithm}

\subsubsection{Test-Time Adaption Version.}
To explore the inter-image identity similarity from unlabeled target data, we exploit the most similar parts among target samples as positive samples to allow modulating the identity distribution.

In generally, gallery images are readily accessible and plentiful when deploying a re-id model in a new scenario, so we do not need to collect extra images and simply sample $N$ gallery images for the test-time adaption.
The part nearest neighbor pairing scheme is proposed to compare the local similarity among $N$ gallery images, and then select $K$ pairs of image with the highest local similarity ($2K \le N$, obviously).
The detailed algorithm is shown in Alg. \ref{alg1}.
It is worth noting that $K$ image pairs are non-overlapping, which means that one image is paired with another \emph{one} at most, thus making maximum use of more samples to stimulate the target distribution more accurately.

The adaption sample set is denoted as $\mathcal{T}=\left\{(\bm{t}_k^n, \bm{t}_k^{n+}) \right\}_{k=1}^K, n \in \left\{h\right\}_{h=1}^H$, where $(\bm{t}_k^n, \bm{t}_k^{n+})$ is a pair of images whose $n$-th body parts are used for test-time adaption.
The test-time adaption version of the part nearest neighbor matching loss $\mathcal{L}_{mat}^{tta}$ is  defined as
\begin{equation}
\label{A_tri}
\mathcal{L}_{mat}^{tta} = \sum_{k=1}^{K} \max(0, \phi + {\rm d}(\bm{f}_k^n, \bm{f}_k^{n+}) - {\rm d}(\bm{f}_k^n, \bm{f}_k^{n-})),
\end{equation}
where ${\rm d}(\bm{f}_k^n, \bm{f}_k^{n+})$ and ${\rm d}(\bm{f}_k^n, \bm{f}_k^{n-})$ are the Euclidean distances between the local feature $\bm{f}_k^h$ and its positive sample $\bm{f}_k^{n+}$ and the hardest negative sample $\bm{f}_k^{n-}$ in a mini-batch, respectively. $\phi$ is the margin parameter.
Similar to the hardest triple loss \cite{triplet}, this loss pulls the local features with the high similarity closer to each other while pushing those with the low similarity away.
By resorting to the inter-image local similarity,
$\mathcal{L}_{mat}^{tta}$ drives the fine-tuning of the identity-aware target distribution. 

\subsection{Training}
At the training phase, \textbf{all} the parameters of our model are optimized by a FSL loss and two SSL losses end-to-end. The training loss $\mathcal{L}_{t}$ and optimization scheme are formulated as
\begin{equation}
\label{training_loss}
\mathcal{L}_{t} = \mathcal{L}_{id} + \lambda_1 \cdot \mathcal{L}_{pos} + \lambda_2 \cdot \mathcal{L}_{mat}^{t},
\end{equation}
\begin{equation}
\theta_*^{all} \leftarrow \theta_*^{all} - \eta_t\nabla_{\theta_*^{all}} \mathcal{L}_{t},
\end{equation}
where $* \in \left\{ \mathcal{E}, \mathcal{G}, \mathcal{P}, \mathcal{M} \right\}$, $\theta_*^{all}$ is all the parameters of $*$.
$\eta_t$ is the learning rate.
$\lambda_1$ and $\lambda_2$ are weighting factors.
The joint learning makes it possible to resort to adjusting the domain-aware structure and identity information for modulating the global feature distribution on the target domain, through associating the three distributions with each other.

\subsection{Batch Norm Test-Time Adaption}
BN is formulated as
\vspace{-2.5mm}
\begin{flalign}
\label{bn_1}
    && & \hat{\bm{x}_b} = \frac{\bm{x}_b - \mu}{\sqrt{\sigma^2 + \epsilon}}, &  \\
    && & \bm{y}_b = \gamma \hat{\bm{x}_b} + \beta, &
\label{bn_2}
\end{flalign}
where $\bm{x}_b$ is the input on the dimension $b$ to a BN layer. $\mu$ and $\sigma^2$ are the empirical mean and variance of the random variable $\bm{x}_b$, which are estimated with a batch of training samples. $\gamma$ and $\eta$ are learnable affine parameters used for linear transformation.
Our experiments in Section \ref{ablation_study} suggest all of the four parameters are biased to the training data to different degrees, leading to the large feature distribution shift between domains.
To this end, two SSL losses are used to update \textbf{BN} parameters for adapting them to the target domain during test-time adaption.
We express the test-time adaption loss $\mathcal{L}_{tta}$ and optimization scheme as
\begin{equation}
\label{test_loss}
\mathcal{L}_{tta} = \mathcal{L}_{pos} + \lambda_3 \cdot \mathcal{L}_{mat}^{tta},
\end{equation}
\begin{equation}
\theta_*^{bn} \leftarrow \theta_*^{bn} - \eta_{tta}\nabla_{\theta_*^{bn}} \mathcal{L}_{tta},
\end{equation}
where $* \in \left\{ \mathcal{E}, \mathcal{P}, \mathcal{M} \right\}$, $\theta_*^{bn}$ indicate BN parameters including the statistics $\mu$, $\sigma^2$ and affine parameters $\gamma$ and $\beta$.
$\eta_{tta}$ is the learning rate and $\lambda_3$ is a weighting factor.
Through fine-tuning parameters, BN layers re-regularize the feature distribution and pull it towards a more stable distribution that can be better processed by the following layers, thereby improving the generalization performance.

\section{Experiments}
\subsection{Datasets and Settings}
\subsubsection{Datasets.}
Following \cite{dualnorm}, we construct the training set by mixing five source domains: CUHK02 \cite{cuhk02}, CUHK03 \cite{cuhk03}, Market-1501 \cite{market1501}, DukeMTMC-ReID \cite{Duke} and CUHK-SYSU PersonSearch \cite{PersonSearch}. All images in the source domains are used for training regardless of train or test splits, covering 121, 765 images of 18, 530 identities in total.
The test sets include VIPeR \cite{viper}, GRID \cite{grid} and iLIDS \cite{ilids}.
Some statistics are listed in Table \ref{testset}.

\subsubsection{Evaluation Protocol.}
We follow the common evaluation metrics for re-id, \emph{i.e.},  mean average precision (mAP) and cumulative matching characteristic (CMC) at rank 1, 5 and 10.
For all the test sets, the average results over 10 random splits are reported.

\subsubsection{Implementation Details.}
Our model is pre-trained on ImageNet \cite{imagenet}, and then trained on the training set for 60 epochs with the Adam optimizer \cite{adam} ($\beta_1$=0.9 and $\beta_2$=0.999).
The learning rate $\eta_t$ is initialized at 0.005, and decayed by 10 after 40 epochs.
The batch size is set to 64.
During test-time adaption, we perform the part nearest neighbor pairing among all the gallery images, which means $N$=316, 900 and 60 for VIPeR, GRID and iLIDS, respectively.
The number of the selected image pairs $K$ is set as $K$=$\min$$\left\{128, \frac{N}{2}\right\}$.
We update our model for only 1 epoch, and randomly sample 32 pairs of images in each batch.
Other hyper-parameters are set as follows:
the number of stripes $H$=6, the weight factor $\lambda_1$=$\lambda_2$=0.1, $\lambda_3$=1, the learning rate $\eta_{tta}$=0.0005, the margin $\phi$ =0.3, the dimension $C$=2048, $C_l$=256.
All the experiments are conducted on a single NVIDIA Titan Xp GPU with Pytorch.

\begin{table}[t]
\centering
\resizebox{.74\columnwidth}{!}{
\begin{tabular}{c|cc|cc}
\toprule[1pt]
\multirow{2}{*}{Datasets} & \multicolumn{2}{c|}{Test IDs} & \multicolumn{2}{c}{Test images} \\
                          & Probe      & Gallery     & Probe       & Gallery      \\ \midrule[0.5pt]
VIPeR                     & 316        & 316         & 316         & 316          \\
GRID                      & 125        & 900         & 125         & 900          \\
iLIDS                     & 60         & 60          & 60          & 60          \\ \bottomrule[1pt]
\end{tabular}
}
\vspace{-2mm}
\caption{Statistics of test sets.}
\vspace{-4mm}
\label{testset}
\end{table}

\subsection{Comparison with state-of-the-art methods}
\begin{table*}[t]
\centering
\resizebox{\textwidth}{!}{
\begin{tabular}{c|cccc|cccc|cccc}
\toprule[1pt]
\multirow{2}{*}{Method} & \multicolumn{4}{c|}{VIPeR}       & \multicolumn{4}{c|}{GRID}        & \multicolumn{4}{c}{iLIDS}           \\
                        & mAP  & R-1 & R-5 & R-10 & mAP  & R-1 & R-5 & R-10 & mAP  & R-1 & R-5 & R-10  \\ \midrule[0.5pt]
DIMN \cite{dimn}                  & 60.1 & 51.2   & 70.2   & 76.0    & 41.1 & 29.3   & 53.3   & 65.8    & 78.4 & 70.2   & 89.7   & 94.5     \\
AugMining \cite{augming}             & -    & 49.8   & 70.8   & 77.0    & -    & 46.6   & 67.5   & 76.1    & -    & 76.3   & 93.0   & 95.3      \\
DualNorm \cite{dualnorm}               & 58.0 & 53.9   & 62.5   & 75.3    & 45.7 & 41.4   & 47.4   & 64.7    & 78.5 & 74.8   & 82.0   & 91.5     \\
BoT \cite{bot}               & 56.7 & 48.2   & -   & -    & 49.6 & 40.5   & -   & -    & 81.3 & 74.7   & -   & -     \\
DDAN \cite{ddan}     & 56.4 & 52.3   & 60.6   & 71.8    & 55.7 & 50.6   & 62.1   & 73.8    & 81.5 & 78.5   & 85.3   & 92.5     \\
DMG-Net \cite{dmg-net}     & 60.4 & 53.9   & -   & -    & 56.6 & 51.0   & -   & -    & 83.9 & 79.3   & -   & -     \\
MetaBIN \cite{metabin}                & 66.0 & 56.9   & 76.7   & 82.0    & 58.1 & 49.7   & 67.6   & 76.8    & 85.5 & 79.7   & 93.3   & 97.3      \\ \midrule[0.5pt]
Baseline                & 60.4 & 50.9   & 72.3   & 80.0    & 48.3 & 38.5   & 57.6   & 65.0    & 82.0 & 76.1   & 89.5   & 94.3      \\
BNTA w/o TTA            & 62.1 & 52.5   & 72.8   & 80.1    & 52.3 & 42.4   & 64.0   & 70.9    & 84.2 & 78.9   & 91.2   & 96.2      \\
BNTA (Ours)             & \textbf{67.3} & \textbf{57.4}   & \textbf{77.6}   & \textbf{82.2}    & \textbf{58.7} & \textbf{51.1}   & \textbf{68.5}   & \textbf{77.3}    & \textbf{85.8} & \textbf{80.6}   & \textbf{93.6}   & \textbf{97.7}    \\ \bottomrule[1pt]
\end{tabular}
}
\vspace{-2mm}
\caption{Comparison with the state-of-the-art generalizable re-id methods (\%). The best results are indicted in bold.}
\vspace{-2mm}
\label{SOTA}
\end{table*}

We compare our model with the existing generalizable re-id methods, including DIMN \cite{dimn}, AugMining \cite{augming}, DualNorm \cite{dualnorm}, BoT \cite{bot}, DDAN \cite{ddan}, DMG-Net \cite{dmg-net} and MetaBIN \cite{metabin}.
The comparison results are displayed in Table \ref{SOTA}, showing that BNTA establishes the new state-of-the-art (SOTA) performance on VIPeR, GRID and iLIDS test sets.
Whether the meta-learning based methods (DMG-Net and MetaBIN), or the methods for domain-invariant learning (DIMN, AugMining and DDAN), or the methods developing BN (DualNorm and BoT), are directly applied to a target domain without being updated.
The superiority of our method over them lies in possessing a domain-adaptive learnability, which capacitates our model to update the model to fit in the target distribution automatically and reduce the distribution shift.



\subsection{Ablation Study}
\label{ablation_study}

\subsubsection{Is BNTA effective?}
To demonstrate the effectiveness of BNTA, we compare it with the baseline and BNTA without test-time adaption (w/o TTA) on three datasets in Table \ref{SOTA}.
The \textbf{baseline} is trained only via FSL, while the \textbf{BNTA w/o TTA} is jointly trained via FSL and SSL, both of which are directly tested on the target domain without being updated.

BNTA w/o TTA achieves 1.6\%, 3.9\% and 2.7\% higher rank 1 scores than the baseline on VIPeR, GRID and iLIDS, respectively.
The performance gain of the additional SSL training results from
facilitating the model to mine local image details for re-id, by predicting the position and learning the similarity for local features.
This phenomenon is similar to the previous findings that exploiting local features boosts the re-id performance \cite{pcb,cvpr2}.

Compared with BNTA w/o TTA, BNTA improves the rank 1 by 4.9\%, 8.7\% and 1.7\% on three datasets, 
which confirms the effectiveness of exploiting body parts to correct the domain bias in BN. 
An explanation about the correlation between body parts and domains is that body parts contain local clothing information, which has different distributions on different domains, due to variation in clothing style, hue, brightness, etc.
The domain shift can be reflected on the change of two self-supervision losses that are built on body parts. 
Then by self-supervised optimization of BNTA, our model can be adapted to the target distribution better.

\begin{table}[t]
\centering
\resizebox{\columnwidth}{!}{
\begin{tabular}{c|cccc}
\toprule[1pt]
\backslashbox{Training}{TTA} & None & $\mathcal{L}_{pos}$ & $\mathcal{L}_{mat}^{tta}$ & $\mathcal{L}_{pos}, \mathcal{L}_{mat}^{tta}$ \\ \midrule[0.5pt]
$\mathcal{L}_{id}$                       & 50.9 & -          & -            & -                      \\
$\mathcal{L}_{pos}$                     & 21.6 & 21.3       & -            & -                      \\
$\mathcal{L}_{mat}^t$                   & 48.2 & -          & 51.8            & -                      \\
$\mathcal{L}_{id}, \mathcal{L}_{pos}$             & 52.6 & 52.8       & -            & -                      \\
$\mathcal{L}_{id}, \mathcal{L}_{mat}^t$           & 52.3 & -       & 55.6         & -                      \\
$\mathcal{L}_{id}, \mathcal{L}_{pos}, \mathcal{L}_{mat}^t$ & 52.8 & 53.3    & 56.8         & \textbf{57.4}                      \\ \bottomrule[1pt]
\end{tabular}
}
\vspace{-2mm}
\caption{Rank 1 of employing different losses during training and TTA on VIPeR (\%). None means not performing TTA.}
\vspace{-3mm}
\label{loss}
\end{table}
\subsubsection{Are both two SSL tasks effective?}
We validate the effectiveness of two SSL auxiliary tasks, \emph{i.e.}, part positioning and part nearest neighbor matching,
by the ablation study of $\mathcal{L}_{pos}$ and $\mathcal{L}_{mat}^t$ (or $\mathcal{L}_{mat}^{tta}$).
As shown in Table \ref{loss}, among multiple combinations of losses, employing $L_{id}$,
\begin{minipage}[t]{\columnwidth}
\hspace{-5mm}
\begin{minipage}[t]{0.48\columnwidth}
\makeatletter\def\@captype{table}
\begin{tabular}{lll}
\toprule[1pt]
Parameter  & Size          & R-1 \\ \midrule[0.5pt]
None           & -          & 42.9 \\ 
Conv          & 93M & 47.8      \\
IN            & 3K  & 43.1      \\
Conv+IN       & 93M & 47.8      \\ \midrule[0.5pt]
BN            & 60K & \textbf{51.1}      \\
BN+Conv       &93M  & 48.1      \\
BN+IN         &93K  & 50.9      \\
BN+Conv+IN    &93M  & 48.0      \\ \bottomrule[1pt]
\end{tabular}
\vspace{-2mm}
\caption{Updating parameters of different layers on GRID (\%).}
\label{different_layers}
\end{minipage}
\hspace{+8mm}
\begin{minipage}[t]{0.44\columnwidth}
\makeatletter\def\@captype{table}
\hspace{+1.5mm}
\begin{tabular}{ll}
\toprule[1pt]
{Parameter}                            & R-1 \\ \midrule[0.5pt]
None                                   & 42.4 \\ 
Statistic $\mu$                        & 44.9      \\
Statistic $\sigma^2$                     & 43.5\\
Statistic $\mu, \sigma^2$                & 45.2     \\ \midrule[0.5pt]
Affine $\gamma$                     & 47.5      \\
Affine $\beta$                      & 46.8      \\
Affine $\gamma, \beta$              & 48.8 \\ \midrule[0.5pt]
All $\mu, \sigma^2, \gamma, \beta$    & \textbf{51.1}      \\ \bottomrule[1pt]
\end{tabular}
\vspace{-2mm}
\caption{Updating different parameters of BN layers on GRID (\%).}
\label{BN_params}
\end{minipage}
\vspace{+3mm}
\end{minipage}
$\mathcal{L}_{pos}$ and $\mathcal{L}_{mat}^t$ for training, $\mathcal{L}_{pos}$ and $\mathcal{L}_{mat}^{tta}$ for test-time
adaption achieves the highest rank 1 score.
Removing $\mathcal{L}_{pos}$ or $\mathcal{L}_{mat}^t$
 (or $\mathcal{L}_{mat}^{tta}$) always decreases the performance in different extents.
The combination of two SSL tasks improves the adaption of the model more obviously, since they help take in the target distribution with respect to the structure and identity information of body parts, respectively.

\begin{figure*}[t]
\centering
\includegraphics[width=\textwidth]{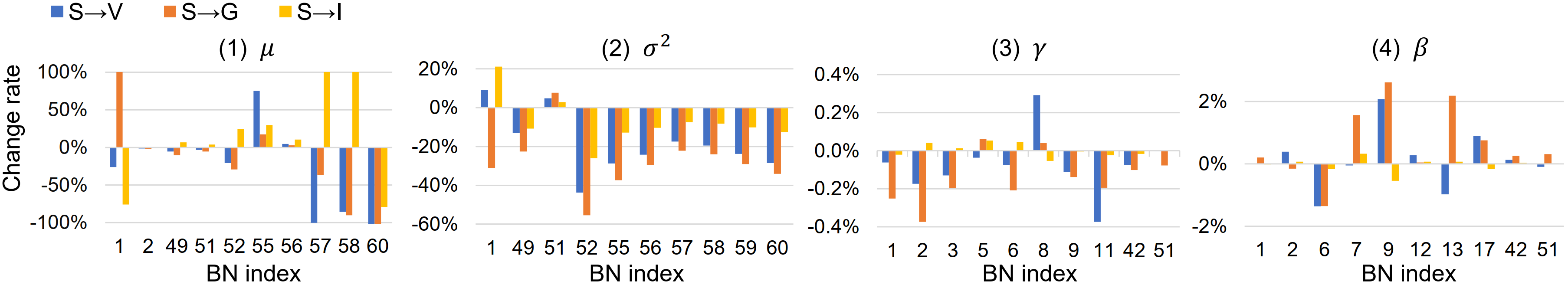} 
\caption{The change rates of BN parameters during test-time adaption. `S$\rightarrow$V', `S$\rightarrow$G' and `S$\rightarrow$I' indicate the transfer from the source dataset to VIPeR, GRID and iLIDS, respectively. For each parameter, the top-10 BN layers that have the largest average change rates in our model are shown here.
BN is indexed according to the order from shallow to deep layers.}
\vspace{-3mm}
\label{bn_vis}
\end{figure*}

\subsubsection{Updating BN or others layers?}
Our model includes three types of layers, \emph{i.e.}, convolution, BN and IN, which are updatable on a new domain.
Table \ref{different_layers} shows the effect of updating different layers during TTA,
 and only updating BN contributes to the best accuracy.
On the one hand, ``BN'' outperforms ``Conv'', ``IN'', ``Conv+IN'' by 3.3\%, 8.0\% and 3.3\% in rank 1, respectively.
This suggests that BN is biased to the training distribution much more seriously than convolution and IN due to the function of normalizing the feature distribution, and is thus more in need of updating on the target distribution.
In fact, whether adding IN (``+IN'') always has quite a small, and even negligible effect on the performance, partly because IN has a much smaller size of parameters than convolution and BN.
Another reason is that IN only regularizes the features over an instance instead of the whole batch of samples like BN, thus not so heavily biased to the distribution of the whole dataset as BN.
On the other hand, ``BN'' achieves the higher accuracy than ``BN+Conv'' and ``BN+Conv+IN''.
This is because the parameter size of convolution (93M) is so larger than BN (60K) and IN (3K) that updating convolution dominates the performance change, and the effect of updating BN is largely weakened.

\begin{figure}[t]
\centering
\includegraphics[width=\columnwidth]{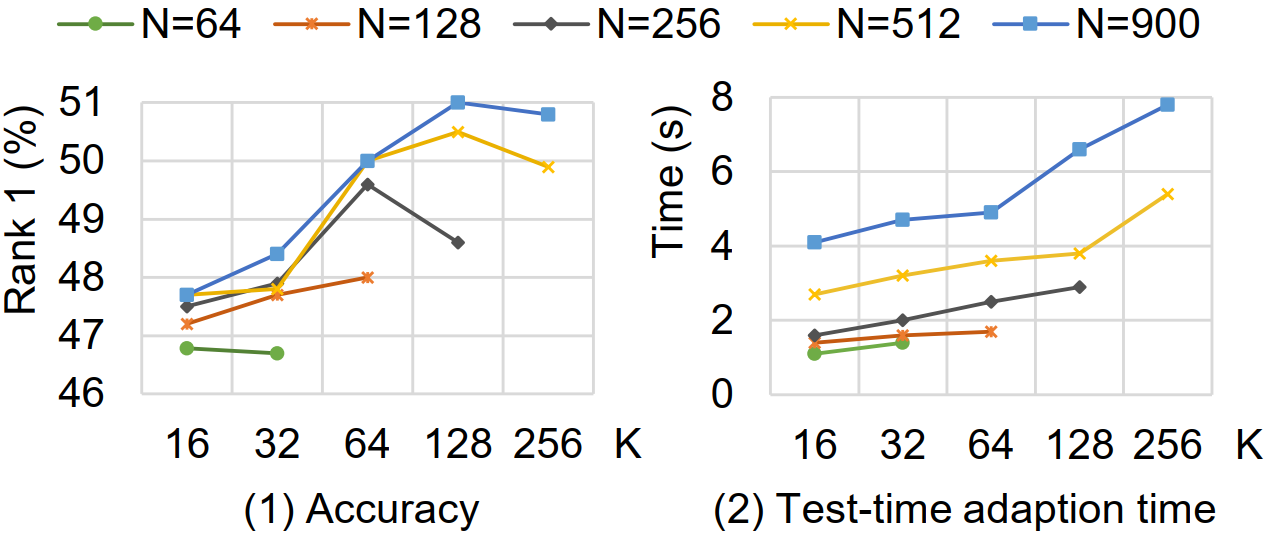} 
\caption{The effect of the hyper-parameter $N$ and $K$ on the accuracy and test-time adaption time on GRID.}
\vspace{-3mm}
\label{N_K_figure}
\end{figure}

\subsubsection{Updating which parameters of BN layers?}
As formulated in Eq. (\ref{bn_1}) and Eq. (\ref{bn_2}), updatable BN parameters include statistics $\mu$, $\sigma^2$ and affine parameters $\gamma$, $\beta$, and Table \ref{BN_params} lists the effects of updating each parameter during adaption.
First, updating any one of the four parameters can boost the performance in different degrees, and the top rank 1 score (51.1\%) is achieved when updating all of them together.
Second, updating affine parameters even brings about more improvements than statistics.
These results manifest that each of the four parameters is closely tied to the domain and suffers from the training bias, but our method can adapt all of them to the target domain and boost the performance fully.

\subsubsection{How much do BN parameters change?}
We illustrate the change rate of each BN parameter during adaption in Fig. \ref{bn_vis}. There are totally 60 BN layers in our model, and we display the top-10 BN layers that have the largest average change rates for each parameter.
The following observations are worth noting.
First, there are significant differences between the change rates of S$\rightarrow$V, G and I, even for the same BN layer and the same parameter.
This reflects various distribution gaps between the source domain and different target domains, and our method can adjust BN parameters to the specific target domain adaptively.
Second, the larger change rate usually takes places at some specific BN layers.
For example, statistics $\mu$ and $\sigma^2$ tend to change more at the BN layers with the index 1, 52, 57, 58 and 60, whereas affine parameters $\mu$ and $\sigma^2$ have larger change rates at 2, 6, 9, 42 and 51, implying these BN layers are more sensitive to the domain shift than others.


\subsubsection{How many samples are required for TTA?}
\label{section_N_K}
$N$ and $K$ are two hyper-parameters controlling the number of samples for TTA.
The part nearest neighbor pairing is performed among $N$ samples, and the top-$K$ pairs with the highest local similarities are selected for updating the model ($2K\le N$).
Fig. \ref{N_K_figure} (1) depicts the effect of the two hyper-parameters, whic exhibits two notable tendencies.
First, given a $N$, the rank 1 tends to first rise and then decline as $K$ increases.
More part nearest neighbors are not always useful and those with the highest similarities can simulate positive samples to facilitate modulating the identity-aware distribution better.
Second, increasing $N$ usually results in the better performance when $K$ is fixed.
A larger $N$ provides more available target samples, so that the top-$K$ pairs of part nearest neighbors can have quite high similarities to serve as positive samples more reasonably.
The performance of our model is likely to be further improved as the number of available target samples grows, which shows the potential of our model in the real-world scenarios that usually allow easy access to a large number of unlabeled samples.

\subsubsection{How much time does TTA cost?}
We show the time cost of the whole TTA process in Fig. \ref{N_K_figure} (2), corresponding to the settings of $N$ and $K$ in Fig. \ref{N_K_figure} (1). 
TTA only takes about 6.8s to update the model before inference to achieve the top rank 1 score ($N$=900, $K$=128). 
On average, inferencing an image on GRID takes 10.7ms and 4.1ms with and without adaption, respectively. 
It is worth noting that the adaption is only performed once, and not needed anymore only if the model is used for inference on the same target domain.

\section{Conclusion}
In this paper, we have proposed a BNTA framework for generalizable re-id, which updates test-time BN layers adaptively on the target domain to correct the training bias carried by BN.
Two part-based SSL auxiliary tasks have been designed to explore the target distribution involved with the structure and identity information within images from unlabeled target samples.
Extensive experiments have shown the effectiveness and potential of updating BN layers for improving the generalization ability.
Only spending a few seconds with hundreds of gallery images for the test-time adaption before inference, our method achieves the state-of-the-art results on three re-id datasets.
In the future work, we will investigate how to update BN and other layers jointly to further enhance the generalization ability.

\section*{Acknowledgements}
This work was jointly supported by National Key Research and Development Program of China Grant No. 2018AAA0100400, National Natural Science Foundation of China (61633021, 61721004, 61806194, U1803261, and 61976132), Beijing Nova Program (Z201100006820079), Shandong Provincial Key Research and Development Program (2019JZZY010119), Key Research Program of Frontier Sciences CAS Grant No.ZDBS-LY-JSC032, and CAS-AIR.

\bibliography{ref}

\end{document}